\documentclass[]{bytedance_seed}

\usepackage[toc,page,header]{appendix}

\usepackage{minitoc}
\usepackage{amssymb}
\usepackage{amsmath}
\usepackage{bm}
\usepackage{graphicx}
\usepackage{subcaption}
\usepackage{array}
\usepackage{wrapfig}

\newcommand{\norm}[1]{\left\lVert#1\right\rVert}

\title{Dexterous Teleoperation of 20-DoF ByteDexter Hand \newline via Human Motion Retargeting
}

\author[]{ByteDance Seed}

\contribution[]{
Full author list in Contributions}

\abstract{
Replicating human-level dexterity remains a fundamental robotics challenge, requiring integrated solutions from mechatronic design to the control of high degree-of-freedom (DoF) robotic hands. While imitation learning shows promise in transferring human dexterity to robots, the efficacy of trained policies relies on the quality of human demonstration data. We bridge this gap with a hand-arm teleoperation system featuring: (1) a 20-DoF linkage-driven anthropomorphic robotic hand  for biomimetic dexterity, and (2) an optimization-based motion retargeting for real-time, high-fidelity reproduction of intricate human hand motions and seamless hand–arm coordination. We validate the system via extensive empirical evaluations, including dexterous in-hand manipulation tasks and a long-horizon task requiring the organization of a cluttered makeup table randomly populated with nine objects. Experimental results demonstrate its intuitive teleoperation interface with real-time control and the ability to generate high-quality demonstration data. Please refer to the accompanying video for further details.
}

\date{\today}
\correspondence{Ruoshi Wen at \email{wenruoshi@bytedance.com}, Zeyu Ren at \email{renzeyu.93@bytedance.com}}

\checkdata[Project Page]{\url{https://byte-dexter.github.io}}

\begin{document}
\maketitle


\section{Introduction}

\begin{figure}[htbp]
    \centering
    \includegraphics[trim=0cm 0cm 0cm 0cm,clip,width=1\textwidth]{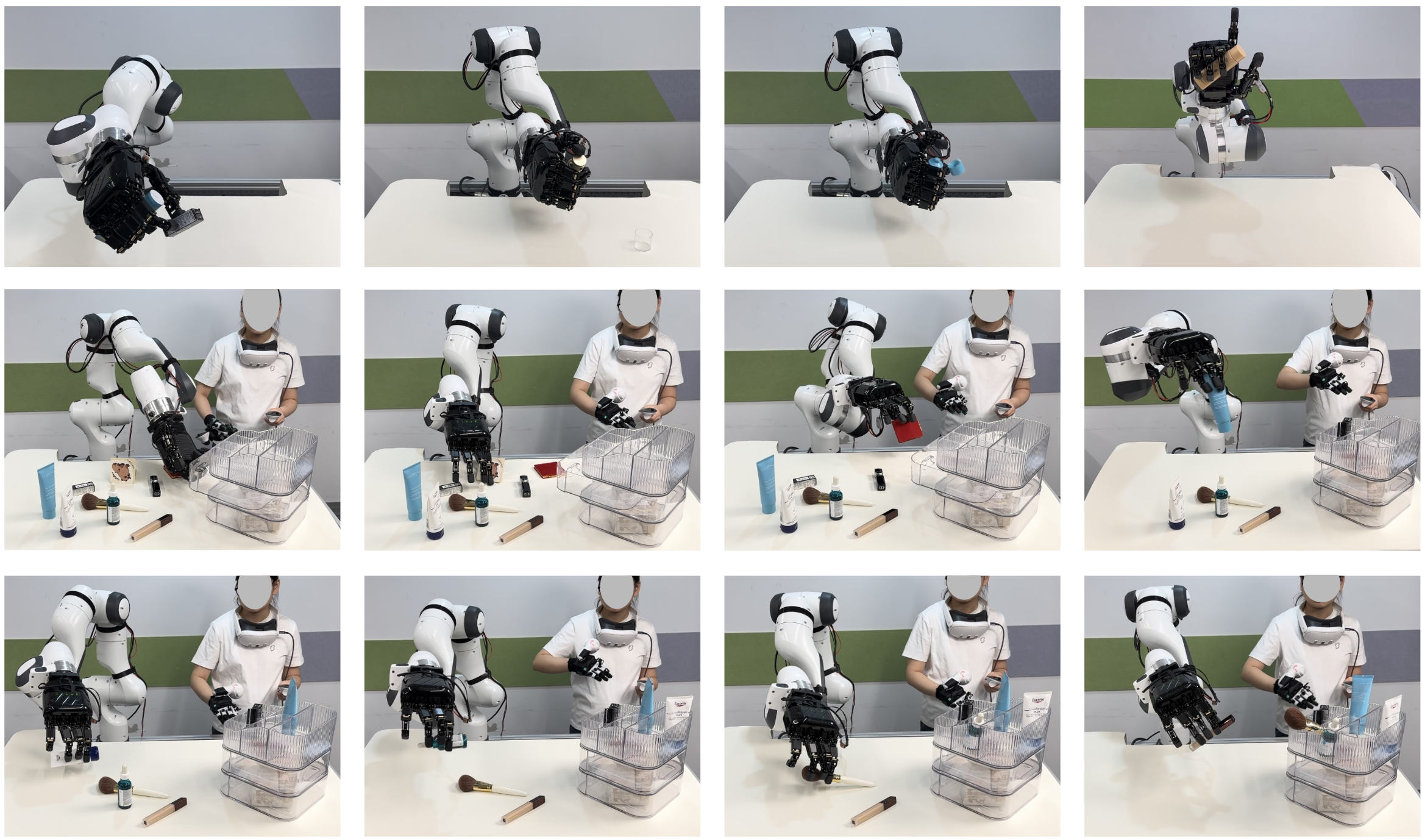}	
	\caption{Our hand-arm teleoperation system achieves dexterous in-hand manipulation, including multi-object grasping, rotation, and regrasping. It also successfully manages grasping of various cosmetic items during a long-horizon cluttered-table organization task.}
    \label{fig:opening}
\end{figure}

As robots transition from structured settings to cluttered, human-centered environments to assist with everyday household tasks, versatile and dexterous manipulation has become critical for widespread, real-world deployment. One promising path to bridge this capability gap for general-purpose robots is to develop anthropomorphic robotic hands that mimic human dexterity through biomimetic design and control~\citep{bicci2019handreview,Han2024handreview}. Developing such systems is a multidisciplinary endeavor that requires the seamless integration of compact, high-degree-of-freedom (DoF) mechanical systems, efficient power-transmission mechanisms, advanced sensing, and precise motor control. Furthermore, achieving human-level dexterity requires control policies capable of coordinating over 20 DoFs to handle complex interactions with objects of diverse shape, material, and mass.

Linkage-driven mechanisms, which transmit motion through cascaded rigid links~\citep{KASHEF2020linkagereview}, offer distinct advantages for compact anthropomorphic hands: durability, ease of maintenance, and actuator integration within the palm. Building on these principles, we present \textbf{ByteDexter}, a 20-DoF linkage-driven anthropomorphic hand featuring: (1) a novel thumb mechanism with three actuators driving four DoFs (abduction–adduction, flexion–extension at the metacarpophalangeal (MCP) joint, flexion–extension at the proximal interphalangeal (PIP) joint, and flexion–extension at the passively coupled distal interphalangeal (DIP) joint), (2) a microsecond-level transmission kinematics solver for real-time control, and (3) a parallel–serial finger topology adapted from~\citep{KimILDA2021}, providing two actuated, coupled DoFs at the MCP joint, one at the PIP joint, and one underactuated DoF at the DIP joint. At a compact form factor (255$\times$118$\times$77 mm$^3$; 1.3 Kg), ByteDexter prioritizes compactness without sacrificing dexterity.

Controlling such high-DoF robotic hands remains challenging. Teleoperation offers a promising path by leveraging human spatial–temporal reasoning to guide robots while generating demonstration data for imitation learning. However, existing systems struggle with kinematic mismatches between human and robotic hands, leading to cognitive operator burden and tedious, error-prone task execution. To address this, we propose an optimization-based motion retargeting framework that minimizes keypoint discrepancies between human and robotic hands, ensuring preserving intentional motions while suppressing involuntary ones that risk collisions or hardware damage.

The contributions of this work include: (1) a 20-DoF linkage-driven hand with a novel thumb mechanics and real-time transmission kinematics solver, (2) a human hand motion retargeting method bridging human-robot kinematic differences for high-fidelity teleoperation, and (3) an integrated hand-arm system enabling dexterous manipulation and seamless coordination across 27 DoFs (20 hand + 7 arm). 

We validate the system in a long-horizon teleoperation task: organizing a cluttered makeup table populated with nine cosmetic items. The system demonstrates robust performance in continuous grasping and manipulation, rapid repositioning to recover from incidental slippage, and execution of in-hand manipulation primitives—including multi-object grasping, rotation and regrasping—achieving human-like dexterity in real-world scenarios.

\section{Related Work}

\label{sec:citations}

\textbf{High-DoF Anthropomorphic Robot Hand} Robotic hands typically employ one of three transmission paradigms: (i) motor-direct-driven actuation~\citep{liu2008hithand,allegro,shaw2023leaphand}, (ii) tendon-driven architectures~\citep{shadow,yang2024trxhand,nathan2024hand}, and (iii) linkage-driven mechanisms~\citep{gros2024hand,KimILDA2021}. Motor-direct-driven mechanism simplifies transmission but ties finger size to actuator dimensions, limiting miniaturization. Tendon-driven hands achieve close biomimicry; however, their reliance on elongated forearms to accommodate tendons, pulleys, and actuators complicates integration with robotic arms, while cable wear necessitates frequent maintenance. Linkage-driven mechanisms prioritize compactness and robustness but face workspace and kinematic complexity tradeoffs. For example, some linkage-based hands adopt simplified three-finger gripper layouts to accommodate cascaded links within constrained palm space~\citep{long2019linkage, Li2024hand}, while the commercial Schunk SVH hand~\citep{schunk} sacrifices dexterity by reducing DoFs per finger to prioritize industrial robustness. Kim et al. advanced linkage-driven dexterity with a modular four-DoF, three-actuator finger design, replicated across five digits to create a 20-DoF anthropomorphic hand~\citep{KimILDA2021}. However, their inverse-kinematics derivation relies on frame-specific trigonometric expansions with numerous intermediate variables, and the absence of an explicit forward-kinematics analysis limits its broader applicability.

\textbf{Robot Hand Teleoperation} Vision-based methods, such as MediaPipe (RGB cameras)~\citep{mediapipe,yang2024ace} and UltraLeap (infrared cameras)~\citep{gil2024dual}, enable markerless tracking of 21 anatomical hand landmarks in real time. However, these approaches remain limited by occlusions, variable lighting, and constrained operational volumes within camera fields of view. While multi-camera setups~\citep{Handa2020} improve tracking accuracy, they introduce logistical complexity. Sensorized gloves bypass these issues by directly measuring joint angles~\citep{udcap} or fingertip poses~\citep{manus,haptx}, but calibration overhead and operator discomfort persist. Despite these tradeoffs, gloves have enabled reliable teleoperation of high-DoF hands like the Allegro and LEAP~\citep{shaw2024bimanual,naughton2024respilot}.

Mapping human poses to robotic joint commands remains challenging due to anatomical mismatches, particularly in thumb kinematics and DoF disparities. Direct one-to-one angle mapping often fails in contact-rich tasks, as misaligned fingers risk collisions or unstable grasps. Recent work increasingly adopts object-centric retargeting to reconcile these differences. One class of methods solves fingertip-level inverse kinematics in the wrist frame \citep{shaw2024bimanual, wang2024dexcap}, using empirical scaling to account for hand-size differences. Other works \citep{naughton2024respilot, Handa2020} formulate it as a constrained-optimization problem that minimizes vector discrepancies between corresponding human and robot keypoints \citep{naughton2024respilot, Handa2020}. Yet these methods struggle to resolve human-robot kinematic differences, compromising teleoperation fidelity for intricate human hand motions.
\section{Robotic Hand-Arm Teleoperation System}
\label{sec:system}

\subsection{System Overview}
\begin{figure}[htbp]
  \centering
  \includegraphics[
    trim={0cm 0cm 0cm 0cm},
    clip,
    width=0.85\textwidth]{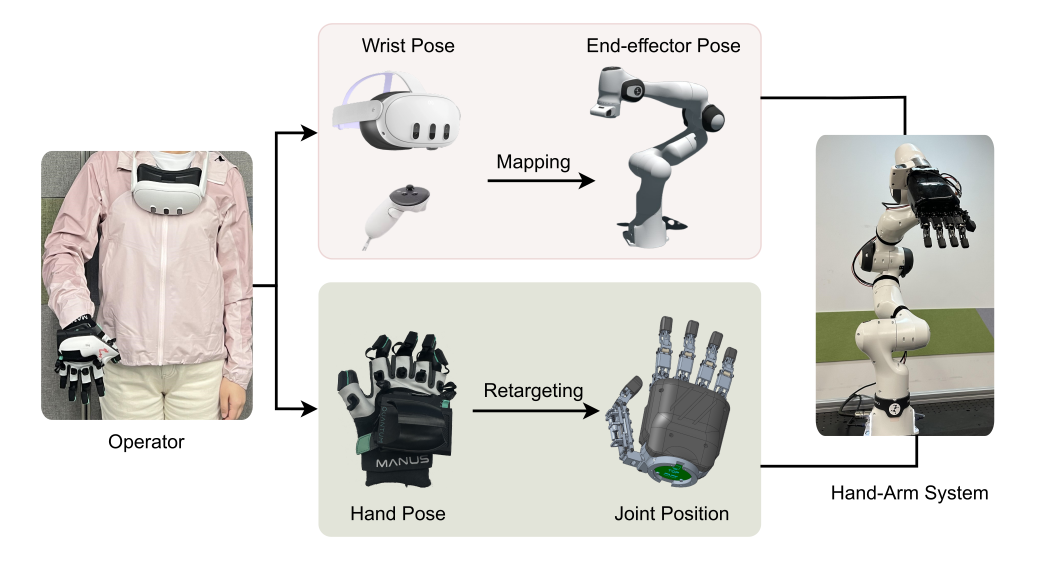}
  \caption{An overview of the proposed hand-arm teleoperation system. The teleoperation interface consists of a Meta Quest 3 and a Manus Quantum Metaglove to track wrist and hand poses simuatenously, and the robotic hand-arm system includes a Franka FR3 arm and the ByteDexter hand.}
  \label{fig:system_overview}
\end{figure}

The hand-arm teleoperation system is illustrated in Figure~\ref{fig:system_overview}. The teleoperation interface comprises a Meta Quest 3 headset to track wrist poses and a Manus Quantum Metaglove for hand motions. The Quest controller is mounted to Manus glove's back via a custom holder to ensure synchronous tracking of wrist and finger movements. This setup delivers an intuitive, natural interface that enhances hand–arm coordination during teleoperation. On the robotic side, the ByteDexter anthropomorphic hand is mounted to a Franka Research 3 (FR3) arm, with its wrist–fingertip axis aligned to the arm’s seventh joint. The operators' wrist poses from the Quest headset are mapped directly to the FR3 end-effector, while hand motions from the Manus Glove are retargeted into joint position commands for the ByteDexter hand.

\subsection{Anthropomorphic Hand System}

\begin{figure}[htbp]
    \centering
    \begin{subfigure}[b]{0.245\textwidth}
        \centering
        \includegraphics[width=\textwidth]{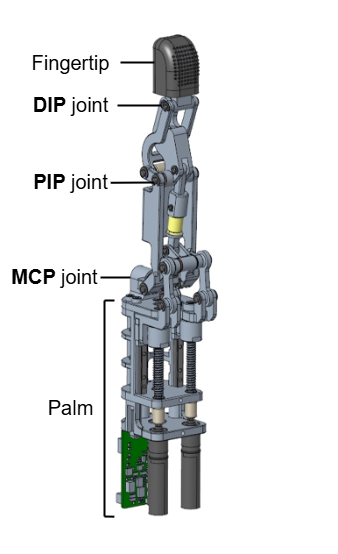}
        \caption{}
        \label{fig:finger}
    \end{subfigure}
    \hfill
    \begin{subfigure}[b]{0.243\textwidth}
        \centering
        \includegraphics[width=\textwidth]{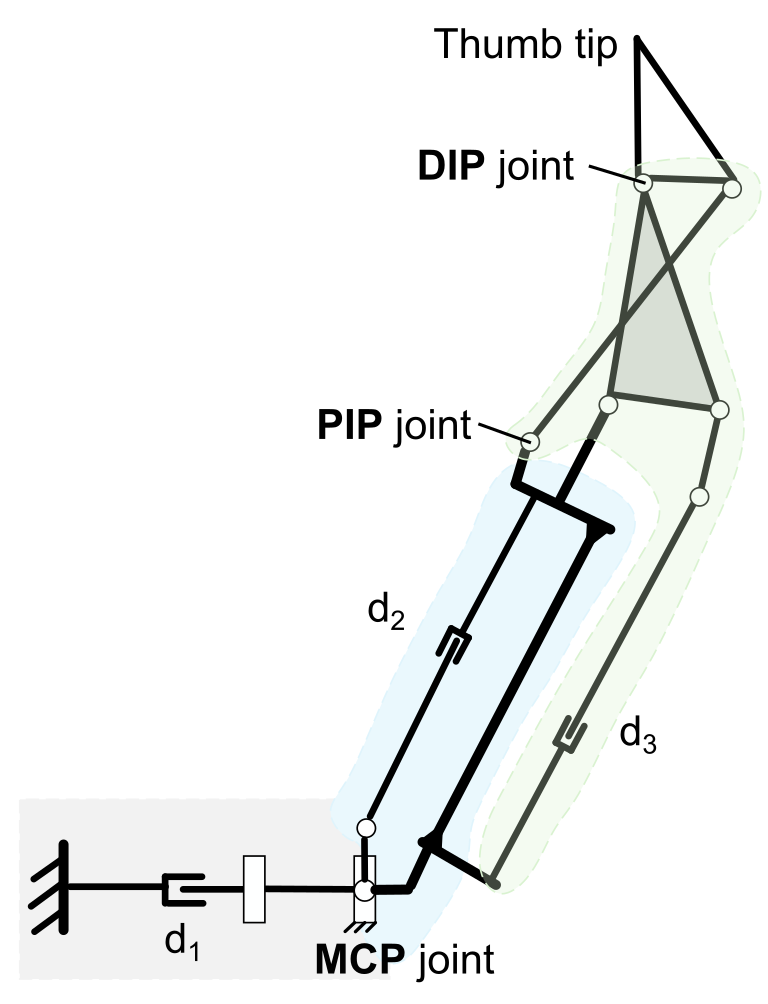}
        \caption{}
        \label{fig:thumb}
    \end{subfigure}
    \hfill
    \begin{subfigure}[b]{0.405\textwidth}
        \centering
        \includegraphics[width=\textwidth]{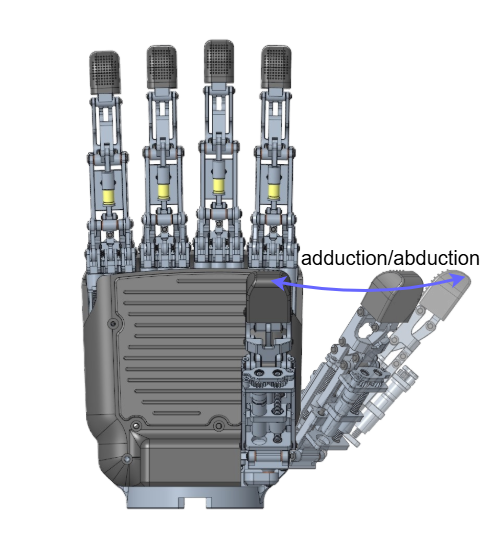}
        \caption{}
        \label{fig:full_hand}
    \end{subfigure}

    \caption{The ByteDexter hand system illustration, showing (a) robotic finger with 2-DoF MCP joint, 1-DoF PIP joint, along with a passively actuated 1-DoF DIP joint, (b) the thumb's kinematic structure, and (c) ByteDexter hand with thumb motion from zero abduction to its full range.}
    \label{fig:hand}
\end{figure}

ByteDexter’s four long fingers (Figure~\ref{fig:finger}) adopt the parallel–serial topology of \citep{KimILDA2021}, but this design proves unsuitable for the thumb. Two DoFs at the MCP joint are coupled by the parallel linkage. As MCP flexion increases, abduction/adduction range diminishes progressively—reaching zero in full flexion. Additionally, implementing thumb abduction/adduction with two actuators mounted perpendicular to the palm unavoidably thickens the palm profile, compromising compactness. To address these limitations, we developed a thumb kinematic structure (Figure~\ref{fig:thumb}) that achieves decoupled, human‐like thumb mobility for advanced grasping and manipulation. Figure~\ref{fig:full_hand} demonstrates the motion range of the proposed thumb topology, allowing full-range MCP and PIP flexion across the thumb's MCP entire abduction (-4$^\circ$ to 90$^\circ$) span.

Alongside ByteDexter's hardware development, we also present a systematic kinematic analysis of the linkage-driven transmission system by formulating both forward and inverse kinematics as sets of constrained nonlinear equations expressed through explicit frame-to-frame transformations (see Supplementary for full derivation). Solving these equations with the Ceres Solver \citep{Agarwal_Ceres_Solver_2022} yields microsecond-level computation times, fast enough to support a 100 Hz control loop for all 15 DoAs. We developed a C++ API that leverages multi-threading for efficient bidirectional communication between the host computer and the onboard controller, computing motor position commands from joint-space commands, and joint states from motor feedback. As joint states are updated each control cycle, we implement a joint-position controller that dynamically adjusts the limit of one DoF at every MCP joint as a function of the other DoF’s position. This ensures that only feasible joint position commands are translated to motor positions and forwarded to the low-level motor drivers.

\subsection{Hand Motion Retargeting}
Manus Gloves use absolute position sensor measurements to track fingertip poses and then estimate joint poses based on human hand models. These sensors are sampled at 120 Hz, and the SDK provides 25 landmarks' poses per glove. We retarget human hand pose data obtained from the Manus Glove into joint position references of the ByteDexter hand by solving an optimization problem that minimizes the difference between corresponding keyvectors in the robotic and human hand.


\begin{wrapfigure}[14]{r}{0.5\textwidth}
  \vspace*{-1\baselineskip}
  \centering
  \includegraphics[width=0.45\textwidth]{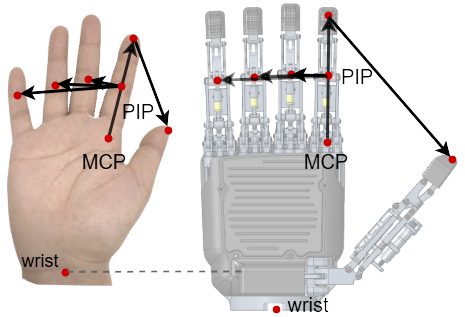}  
  \caption{Hand keypoint vectors.}
  \label{fig:retargeting}
\end{wrapfigure}

A keyvector is the 3D vector between a pair of key points, pointing from the origin of one coordinate frame to the other, expressed in the robot hand's base frame. The finger design of ByteDexter (from the MCP joint to the fingertip, Figure~\ref{fig:finger}) adopts anatomically proportional dimensions scaled to human finger lengths. However, the palm size is constrained to prioritize functional integration of motors, embedded boards, and leadscrews, resulting in a non-anthropomorphic structure that introduces morphological discrepancies with human palm anatomy. To address this non-anatomical factor, our approach diverges from prior methods using fingertip-to-wrist keyvectors; instead, we compute keyvectors from each fingertip to its own MCP joint. For clarity, Figure~\ref{fig:retargeting} shows only the index-finger keyvectors: from the MCP joint to the fingertip, from the fingertip to the thumb tip, from the PIP joint to the PIP joints of the middle and ring fingers, and from PIP joint to the pinky's DIP joint (reflecting the pinky's shorter length). Iterating this process from the thumb to pinky and eliminating duplicates results in 15 unique keyvectors. Incorporating these keyvectors into the optimization captures the majority of the grasping types: for pinch grasping, minimizing inter-finger distances (e.g., thumb-index, adjacent fingers) aligns robotic and human hand poses to reduce operator strain, whereas for power grasping, preserving fingertip-to-palm distances are prioritized.

Hand motion retargeting is formulated as the following optimization problem, minimizing the differences between human hand keyvectors and those on the robotic hand:

\vspace{-3mm}
\begin{equation}\label{eq:obj}
\begin{gathered}
  \min_{q_t}\;\sum_{i=0}^{N} w(d_i)\norm{r_i(q_t)-f(d_i)\hat v_{i,t}}^2
    + \lambda \norm{q_t - q_{t-1}}^2,\\
  \text{s.t.}\quad q_{l} \le q_{t} \le q_{u}\,.  
\end{gathered}
\end{equation}
where $q_t$ is the joint position vector of the robotic hand, $r_i(q_t)$ represents the $i^{th}$ keyvector calculated using forward kinematics on the robotic hand. Furthermore, $d_i=\norm{v_{i,t}}$ and $\hat v_{i,t} = {v_{i,t}\over \norm{v_{i,t}}}$, $v_{i,t}$ is the $i^{th}$ keyvector on the operator's hand. $w(d_i)$ is the weight function of $d_i$, defined as
\[
w(d_i) =
\begin{cases}
1,    & d_i > \epsilon\\
200,  & d_i \le \epsilon \land v_{i,t} \in \mathcal{S}_1 \\
400, & d_i \le \epsilon \land v_{i,t} \in \mathcal{S}_2
\end{cases} \ ,
\] and $f(d_i)$ defines a distance conditioned on $d_i$:
\[
f(d_i) =
\begin{cases}
\beta_i d_i,    & d_i > \epsilon\\
\eta_1,  & d_i \le \epsilon \land v_{i,t} \in \mathcal{S}_1 \\
\eta_2, & d_i \le \epsilon \land v_{i,t} \in \mathcal{S}_2
\end{cases} \ ,
\]
where $\beta_i \in \mathbb{R}^3$ is the scaling factor for the $i^{th}$ keyvector, and $\eta_1$ forces closing the distance between a finger and the thumb, while $\eta_2$ forces a separation distance between two fingers to avoid inter-finger collisions. $\mathcal{S}_1$ denotes the set of vectors that originate from a finger (index, middle, ring, pinky) and point to the thumb, and $\mathcal{S}_2$ is a set of inter-finger vectors. We add the last term to improve temporal smoothness.

Optimizing Equation~(\ref{eq:obj}) requires evaluating each link's pose via forward kinematics from $q_t$, and computing the corresponding gradients. The nonlinear coupling between the underactuated DIP and PIP joints—expressed by the constraint $h(q_{PIP}, q_{DIP})=0$ complicates this Jacobian calculation. We address it by applying the chain rule: first compute $\partial q_{DIP}\over{\partial q_{PIP}}$ from the constraint $h$, then incorporate this term into the PIP column of the overall Jacobian, thus capturing the DIP–PIP dependency.

\subsection{Robotic Arm Teleoperation}

Meta Quest headset captures its controller's transform relative to its start pose, which are mapped as pose references for the FR3 end-effector. Human wrist pose mapping is formulated as a constrained optimization problem (Equation~\ref{eq:arm_obj}) that enforces joint limits, generates smooth motion and natural configurations. The VR teleoperation loop runs at 50 Hz, while the FR3 arm tracks the optimized motions with a 1 kHz joint-impedance controller.

\begin{equation}\label{eq:arm_obj}
\begin{gathered}
\underset{\dot q}{\min}  \left \| J_{e} \dot{q} - \lambda \Delta x \right \| ^2 _{w_{0}} 
+ \left \| J_{a} \dot{q} - \Delta \alpha  \right \| ^2 _{w_{1}} 
+ \left \| {\dot q} \right \| ^2 _{w_{2}} 
+ \left \| {\dot q} - {\dot q}_{t-1} \right \| ^2 _{w_{3}} 
 \\
\text{s.t.}\quad f_{m} \leq   {\dot q} \leq   f_M
\end{gathered}
\end{equation}
where $J_{e}$ and $\Delta x$ represent the tool frame Jacobian and relative transforms of the human wrist poses,  $\lambda$ is the scaling factor for compensating the workspace difference between human and robotic arm, $J_{a}$ and $\Delta \alpha$ are the arm angle Jacobian and deviation from vertical reference plane, $w_{\{i\}}$ is the weights of different tasks, and $f_{m}$, $f_{M}$ represent the constraints of joint position and velocity. We add the last term to avoid sudden acceleration and improve motion smoothness. 

\section{Experimental Results}
\label{sec:result}

\subsection{Experimental Setup}

We first benchmark our retargeting pipeline against a modified DexPilot~\citep{qin2023anyteleop} implementation on thumb–index and thumb–middle pinch motions. For each grasp, we quantify performance by measuring the inter-fingertip distance between the primary digits and monitoring unintended collisions among the remaining fingers.

We then assess our teleoperation framework on a wide range of dexterous grasping and manipulation tasks - spanning precision and power grasps, as well as non-prehensile manipulation, using a cluttered makeup table populated with objects of diverse size, shape, mass, and surface friction. An operator equipped with a Manus Meta Glove and Meta Quest controller, cleared the table in one continuous trial. To further evaluate the retargeting pipeline, we introduced advanced in-hand manipulation scenarios. Demonstrations of these tasks are provided in the accompanying video.

\subsection{Retargeting Comparison Results}

\begin{figure}[htbp]
    \centering
    \begin{subfigure}[b]{0.48\textwidth}
        \centering
        \includegraphics[trim={2cm 1cm 4cm 2cm},
    clip, width=\textwidth]{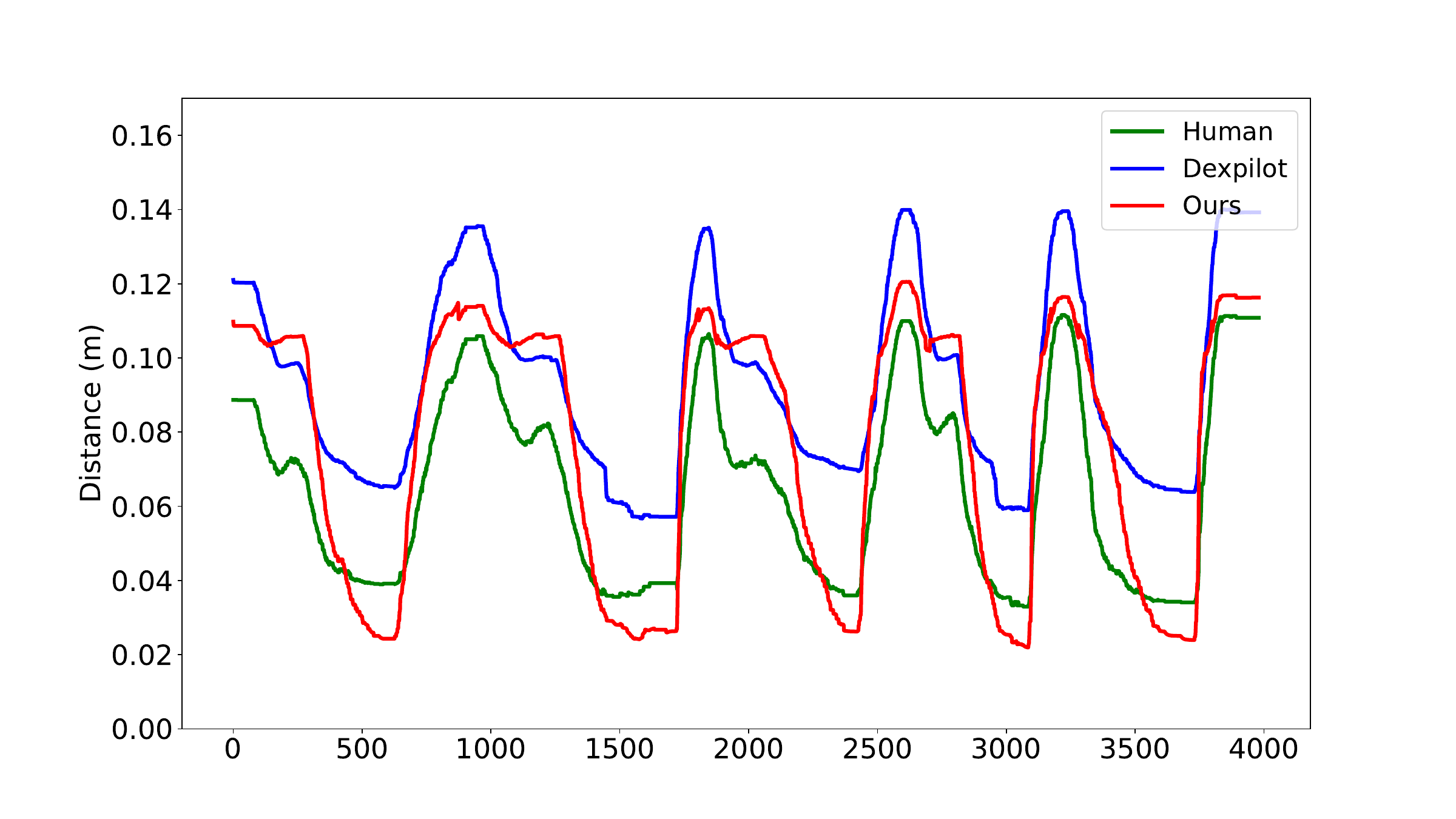}
        \caption{Thumb-index fingertip distances}
        \label{fig:thumb_index}
    \end{subfigure}
    \hfill
    \begin{subfigure}[b]{0.48\textwidth}
        \centering
        \includegraphics[trim={2cm 1cm 4cm 2cm},
    clip, width=\textwidth]{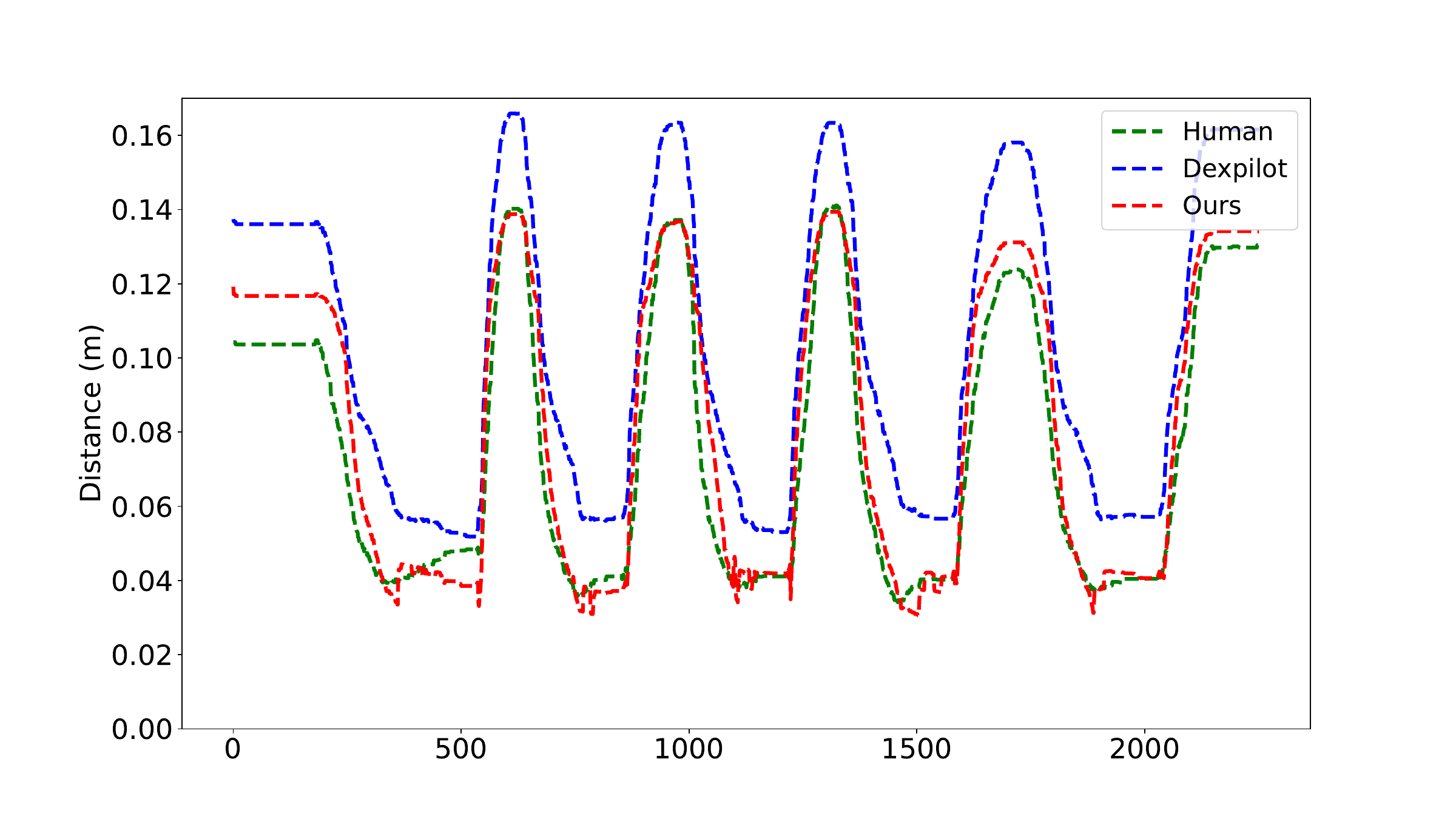}
        \caption{Thumb-middle fingertip distances}
        \label{fig:thumb_middle}
    \end{subfigure}

    \caption{Comparison between our retargeting method and DexPilot, as quantified by the thumb–index and thumb–middle fingertip distances.}
    \label{fig:comparison}
\end{figure}

Figure~\ref{fig:comparison} plots thumb-index and thumb-middle fingertip distances over five open-close pinch cycles, comparing human reference trajectories (green) with robot motions from our retargeting method (red) and DexPilot (blue). While both methods follow the operator’s pinch profile, our approach consistently achieves tighter spacing in thumb–index experiments (Fig.~\ref{fig:thumb_index}). By maintaining a tighter distance than the human reference, the fingertips can exert forces into the object, thus improving grasp stability. In addition to optimizing inter-finger distances to match human references, our retargeting reproduces human-like joint coordination—for example, aligning opposing thumb and index fingertips into a parallel, gripper-like posture—which proved beneficial for reliably retrieving slender items (e.g., lipsticks, brushes, tubes) during cluttered makeup-table cleanup. Furthermore, our method reduced collisions between task-irrelevant fingers compared to DexPilot (Fig.~\ref{fig:thumb_middle}), mitigating risks of mechanical wear during prolonged teleoperation sessions. These combined advantages position our retargeting method as a robust solution for robotic grasping and manipulation.

\subsection{In-hand Manipulation Results}

We evaluate three canonical in-hand manipulation primitives—(i) regrasping to translate objects between fingers, (ii) sliding objects relative to the palm, and (iii) rotating objects or their components—and instantiate them in four benchmark tasks: (1) regrasping a bottle from a precision to a power grasp; (2) multi-object grasping; (3) lid unscrewing; and (4) push-to-open lids.

\begin{figure}[htbp]
    \centering
    \begin{subfigure}[b]{0.95\textwidth}
        \centering
        \includegraphics[trim={0cm 0.8cm 0cm 0cm},
    clip, width=\textwidth]{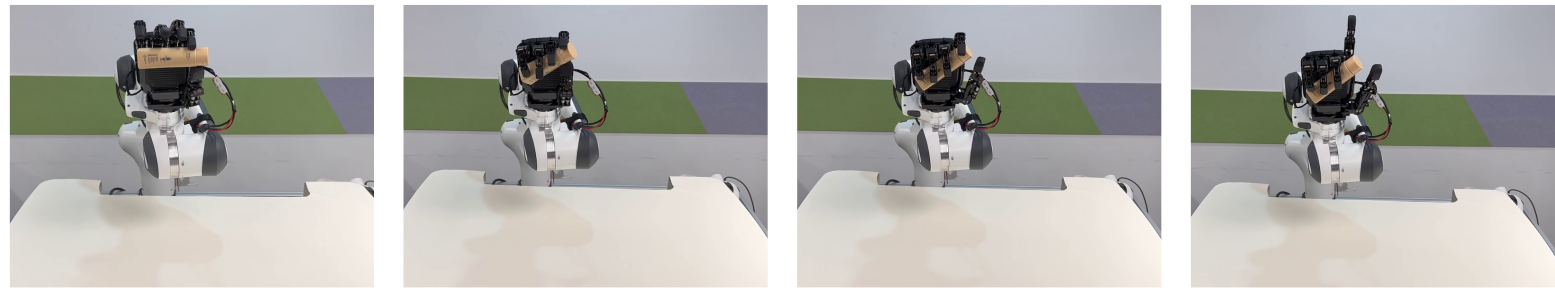}
        \caption{Regrasping via switching grasp type: from thumb-index precision pinch to power grasp}
        \label{fig:in_hand1}
    \end{subfigure}
    \begin{subfigure}[b]{0.95\textwidth}
        \centering
        \includegraphics[width=\textwidth]{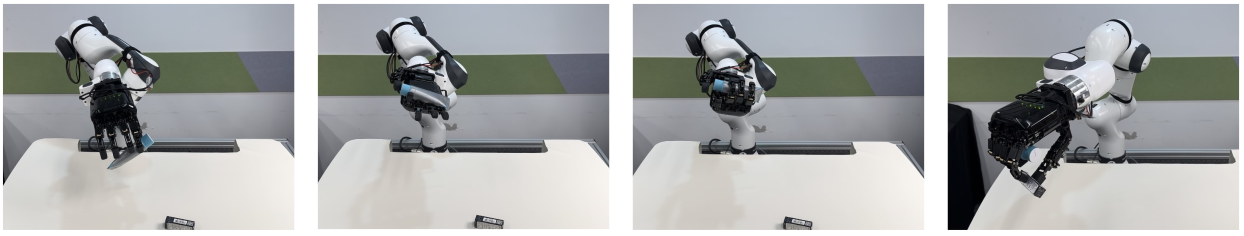}
        \caption{Simultaneous grasping of two objects through shifting the first object along the palm}
        \label{fig:in_hand2}
    \end{subfigure}
     \begin{subfigure}[b]{0.95\textwidth}
        \centering
        \includegraphics[trim={0cm 0.3cm 0cm 0cm},
    clip, width=\textwidth]{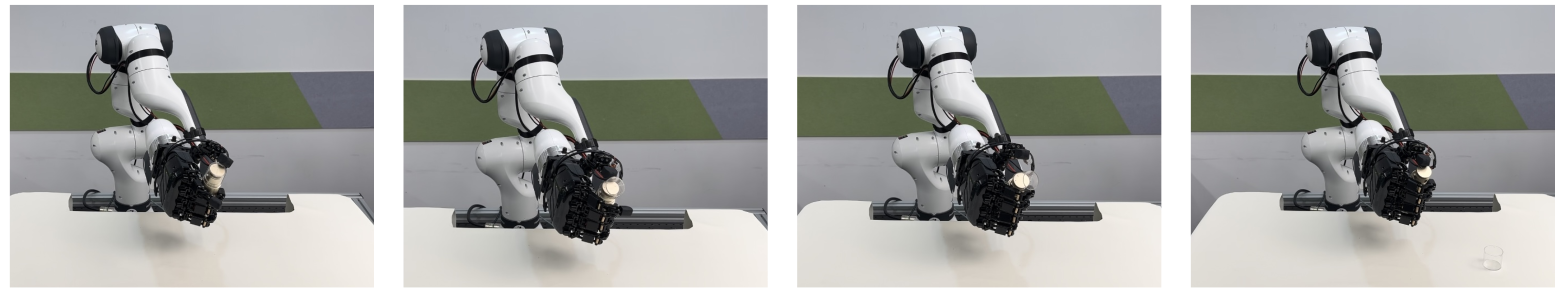}
        \caption{Lid opening via pushing}
        \label{fig:in_hand3}
    \end{subfigure}
    \begin{subfigure}[b]{0.95\textwidth}
        \centering
        \includegraphics[trim={0cm 0.3cm 0cm 0cm},
    clip, width=\textwidth]{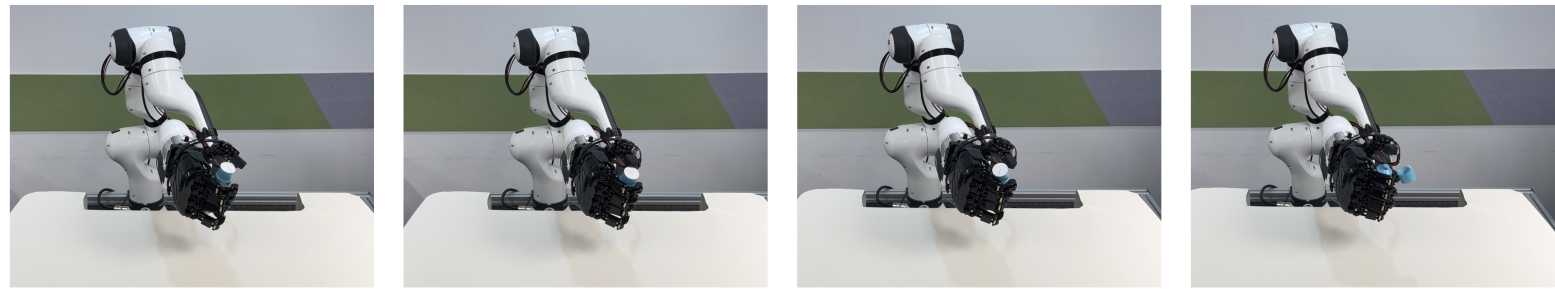}
        \caption{Lid opening via rotating}
        \label{fig:in_hand4}
    \end{subfigure}
    \caption{In-hand manipulation tasks.}
    \label{fig:horizontal_images}
    \vspace{-5mm}
\end{figure}

Regrasping an object (shown in Figure~\ref{fig:in_hand1}) involves switching contacts with the hand to achieve an optimal grasp configuration. Typically, a thumb-index precision grasp is employed to retrieve the object from a surface, followed by a transition to a power grasp that engages the remaining fingers and the palm to stabilize the hold. This regrasping strategy leverages the accuracy of precision grasping for initial acquisition and the stability of power grasping for secure manipulation, thereby enhancing grasp reliability while reducing the physical and cognitive burden on the operator. 

As illustrated in Figure ~\ref{fig:in_hand2}, the operator simultaneously grasps two objects by sequentially reconfiguring the hand. First, the operator retrieves a cream tube from the table with the thumb–index precision grasp. To free the thumb and index finger for a second grasp, the pinch is switched to a power grasp that engages the middle, ring, and pinky fingers together with the palm. After the retrieval of the first object, the operator rotates the hand palm-up to serve as a holder when pushing the tube along the palm until the index finger is clear. The hand is then rotated palm-down, the arm is moved toward a second object on the table, and a thumb-index precision grasp is executed to retrieve that object. This grasping strategy allows the operator to hold multiple objects concurrently, demonstrating a previously unseen level of dexterity.

Lid opening represents a critical category of in-hand manipulation, as it requires manipulating the lid while maintaining control over the main body of the object. We evaluated two distinct lid-opening strategies: unscrewing via rotational motion and pushing using the thumb (see Figure~\ref{fig:in_hand3} and~\ref{fig:in_hand4}). In both cases, the applied force or torque must be carefully regulated—sufficient to initiate lid movement, yet not so large as to compromise the stability of the grasp on the main body.

\subsection{Long-Horizon Teleoperation Results}

We evaluated the system’s long-horizon teleoperation capabilities through a table organization task involving a cluttered workspace populated with randomly arranged cosmetic and skincare items, including serum bottles, cream tubes, lipsticks, brushes, and powder palettes. A multi-drawer makeup organizer was placed adjacent to the clutter, requiring the robotic hand-arm system to retrieve objects and place these in the organizer, additionally, insert items into a drawer. As illustrated in Figure~\ref{fig:opening} (bottom rows), the system successfully managed diverse geometries and recovered from grasp slippage through real-time adjustments, demonstrating robust performance in unstructured environments.

Drawer opening presents a significant challenge: the recessed finger pull at the bottom of the drawer is roughly the same size as a fingertip and lies near the tabletop, constrained by the arm’s workspace. To engage it, the operator tilts the hand about 45° upward, flexes the index finger to insert the fingertip vertically into the recess, and then retracts the arm to pull the drawer open. When pushing to close the drawer, it is found that distributing the pushing force across multiple fingertips could even out the resulting torque which prevents the entire organizer from tilting - a problem that arises when the force is applied with one fingertip.

We further demonstrate that non‐prehensile manipulation primitives—pushing, pulling, and tilting—substantially improve task efficiency when clearing a cluttered makeup table. For example, a red pocket mirror just 5 mm thick is first dragged to the table’s edge to expose its narrow profile, then secured with a thumb–index pinch grasp. To grasp a beige‐colored palette, the hand pushes against its edge to flip it, thereby presenting a larger surface for a stable grip. We also validate precision grasps on slender items: in one trial, the hand closes a lipstick cap without disturbing its pose, underscoring the system’s next‐level dexterity.

Our hand-arm teleoperation system delivers dexterity and flexibility, enabling fast object repositioning and reliable re-grasping. In a nine-object sequence—including opening and closing a drawer—the operator completed the task in under five minutes despite occasional slippage. With further operator training, we anticipate even faster completion times, making our platform a robust solution for continuous data collection in imitation-learning pipelines and the eventual automation of such tasks.

\section{Conclusion}

\label{sec:conclusion}

We present a hand-arm teleoperation system featuring a 20-DoF linkage-driven robotic hand and a hand motion retargeting method. The glove-based operation interface enables seamless hand-arm coordination and the retargeting approach achieves precise human hand motion reproduction. The real-world grasping and manipulation experiments' results have demonstrated the potential of collecting high-fidelity data for imitation learning. In the future, we would like to investigate bimanual teleoperation system and automating the demonstrated dexterous skills.

\section{Limitations}
\label{sec:limit}

Human thumbs benefit from the saddle joint to achieve true oppositional with each of the rest four fingers. This capability is not yet realized by the ByteDexter hand. While the current thumb–index pinch successfully enables precision grasping across a range of objects, the absence of full opposition limits (1) the variety of manipulable geometries (e.g., irregular or deformable items) and (2) the execution of seamless finger-gaiting manipulation. Developing a thumb that has this oppositional function would expand the thumb’s workspace, promote more balanced force distribution and optimal contact surfaces, and thereby enhance grasp stability in complex in-hand manipulation tasks.

While the Meta Quest 3 delivers robust wrist‐pose estimation under nominal conditions, occlusions remain a key limitation: if the Quest controller is rotated out of the headset’s line of sight, tracking is lost—an issue familiar to all vision‐based systems. Although our mapping pipeline mitigates sudden jumps by holding the arm’s last known command until the Quest controller reappears, long‐term reliability will require sensor fusion with complementary modalities (e.g., inertial measurement units or alternative wearable trackers) to maintain seamless arm tracking even under full occlusion.

Long-term teleoperation reveals a critical limitation: human operators experience significant cognitive fatigue when managing low-level grasp stabilization and high-level task planning simultaneously. While operators excel at scene comprehension, grasp-type selection, and sequencing long-horizon actions, the constant need to manually regulate contact forces and joint-level stability diverts attention from strategic decision-making. This imbalance underscores the necessity for future integration of autonomous low-level grasping strategies—such as closed-loop force modulation to offload repetitive stabilization efforts. By offloading these low-level subtasks to the system, operators can concentrate on high-level decisions—such as what and how to grasp—thereby reducing fatigue, improving task efficiency, and enhancing data quality during prolonged data-collection sessions.

\clearpage
\section{Contributions}
\linespread{1.5}\selectfont 
\textbf{ByteDexter Design} 
\newline
Jiajun Zhang, Zhigang Han, Hao Niu, Min Du, Junkai Hu, Yang Gou, Guangzeng Chen, Zeyu Ren

\textbf{Control and System} 
\newline
Ruoshi Wen, Zhongren Cui, Liqun Huang, Haoxiang Zhang, Zhengming Zhu, Wei Xu

\textbf{Writing and illustration} 
\newline
Ruoshi Wen, Liqun Huang, Jiajun Zhang, Wei Xu, Zeyu Ren

\textbf{Project Leader} 
\newline
Zeyu Ren, Hang Li

\textbf{Affiliation} 
\newline
All contributors are from ByteDance Seed.

\textbf{Acknowledgment} 
\newline
We would like to sincerely thank Xiao Zhang and Mingyu Lei for their support in setting up the experiments. 
\clearpage

\bibliographystyle{plainnat}
\bibliography{main}

\clearpage

\beginappendix

\section{Kinematic Analysis of the Finger}

\begin{figure}[htbp]
    \centering
    \begin{subfigure}[b]{0.3\textwidth}
        \centering
        \includegraphics[width=\textwidth]{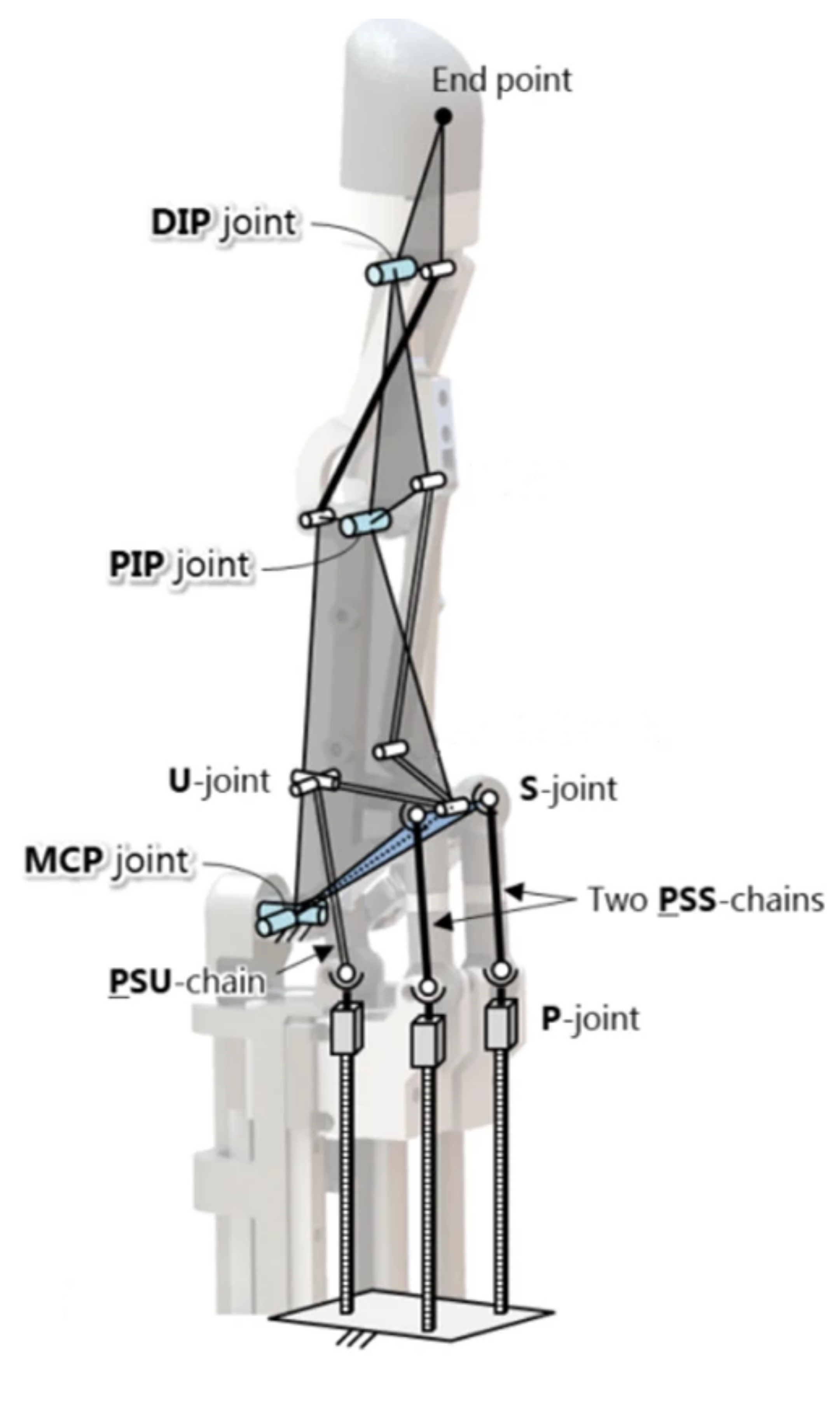}
        \caption{}
        \label{fig:finger_chain}
    \end{subfigure}
    \hfill
    \begin{subfigure}[b]{0.67\textwidth}
        \centering
        \includegraphics[width=\textwidth]{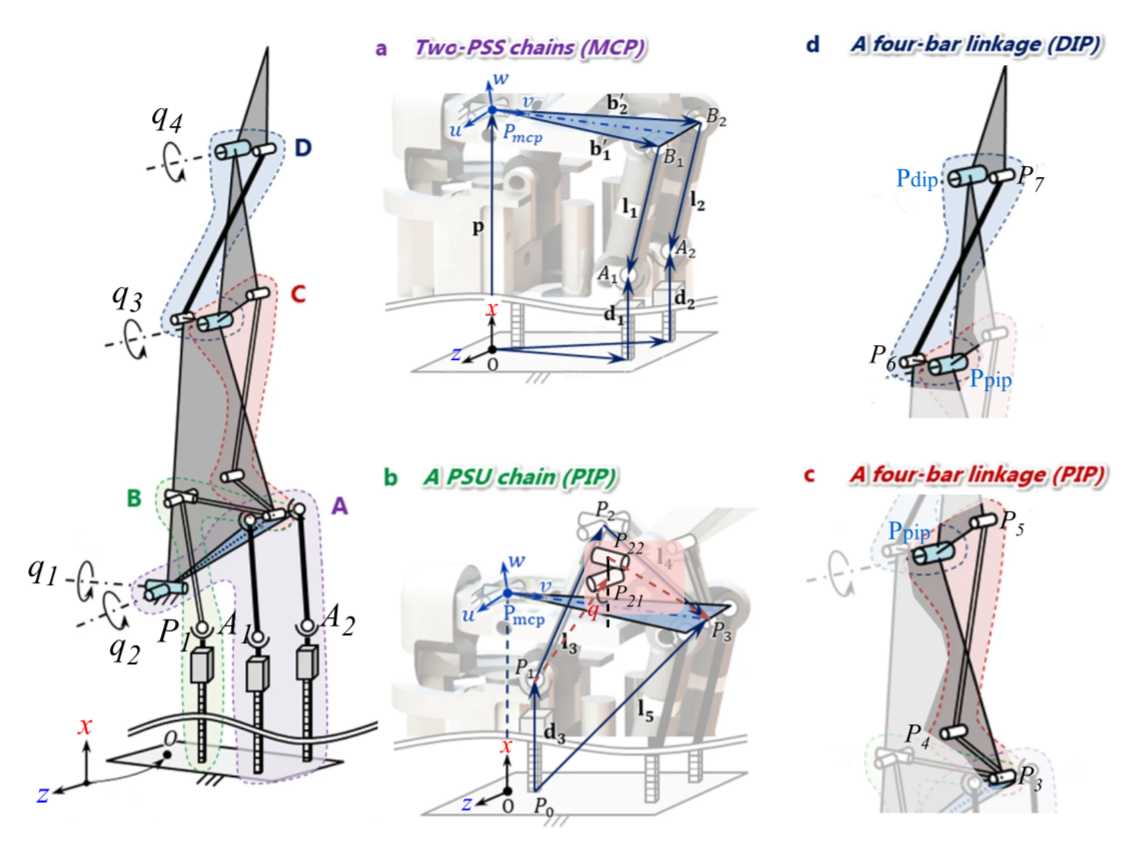}
        \caption{}
        \label{fig:finger_kinematics}
    \end{subfigure}

    \caption{Kinematic analysis of the ByteDexter finger. (a) Kinematic chains of MCP, PIP and DIP. (b) The kinematic structure of the finger.\\ Adapted from Kim et al., \textit{Nature Communications} (2021) \url{https://doi.org/10.1038/s41467-021-27261-0}, licensed under \href{https://creativecommons.org/licenses/by/4.0/}{CC BY 4.0}.}
    \label{fig:finger_analysis}
\end{figure}

The kinematic structure of the finger is depicted in Figure~\ref{fig:finger_analysis}. The rotary motions of three motors are transmitted into linear motions at three spherical joints via lead screws, which are represented as prismatic joints. As shown in Figure~\ref{fig:finger_chain}, two prismatic-spherical-spherical (PSS) chains drive the 2-DoF motion of the MCP joint, enabling abduction/adduction and flexion/extension. The PIP joint is actuated by a prismatic-spherical-universal (PSU) chain combined with a crossed four-bar linkage, and the DIP joint is coupled to the PIP joint through an additional crossed four-bar linkage.

In the following sections, we analyze the kinematics of these mechanisms, including the two PSS chains of the MCP joint, the PSU chain and four-bar linkage of the PIP joint, and the four-bar linkage of the DIP joint. We define a global coordinate frame, $O-xyz$ where the origin $O$ is the projection of the MCP joint onto the plane formed by the endpoints of the three P joints. Notably, we represent the transformation of frame $B$ relative to frame $A$ as $(\mathbf{R}^A_B, \mathbf{p}^A_B)$, where $\mathbf{R}^A_B$ is the rotation matrix that describes the orientation of frame $B$ with respect to frame $A$, and $\mathbf{p}^A_B$ is the position vector from the origin of frame $A$ to the origin of frame $B$, expressed in frame $A$.

\subsection{MCP Kinematics}

\textbf{Two PSS-Chains for the 2-DoF Motion} At the MCP joint, two rotations ($q_1$, $q_2$) enable abduction/adduction and flexion/extension motions (see Figure~\ref{fig:finger_kinematics}-a). The relationship between $\mathbf{q}$ and $\mathbf{d}$ is denoted as $\boldsymbol{f}(\mathbf{q}, \mathbf{d}) = \mathbf{0}$, where $\mathbf{q} = [q_1, q_2]^T$, $\mathbf{d} = [d_1, d_2]^T$, and $\boldsymbol{f} = [f_1, f_2]^T$. $\boldsymbol{f}$ is derived as follows:

\begin{equation}\label{eq:mcp}
\begin{gathered}
  \mathbf{R}^{O}_{P_{mcp}} = \mathbf{R}_y(q_1) \mathbf{R}_z(q_2),\\
  \mathbf{p}^O_{B_1} = \mathbf{p}^O_{P_{mcp}} + \mathbf{R}^O_{P_{mcp}} \mathbf{p}^{P_{mcp}}_{B_1}, \ \ \mathbf{p}^O_{B_2} = \mathbf{p}^O_{P_{mcp}} + \mathbf{R}^O_{P_{mcp}} \mathbf{p}^{P_{mcp}}_{B_2}, \\
  \mathbf{p}^O_{A_1} = \mathbf{p}^O_{A_{10}} - [d_1, 0, 0]^T, \ \ \mathbf{p}^O_{A_2} = \mathbf{p}^O_{A_{20}} - [d_2, 0, 0]^T, \\
  ||\mathbf{p}^O_{B_1}-\mathbf{p}^O_{A_1} || = ||\overrightarrow{A_1B_1}|| =  l_1, \ \ ||\mathbf{p}^O_{B_2}-\mathbf{p}^O_{A_2} || = ||\overrightarrow{A_2B_2}|| = l_2,
\end{gathered}
\end{equation}
where $\mathbf{p}^O_{A_{10}}$, $\mathbf{p}^O_{A_{20}}$, $\mathbf{p}^O_{B_{10}}$, and $\mathbf{p}^O_{B_{20}}$ denote the initial position vectors of points $A_1$, $A_2$, $B_1$, and $B_2$, expressed in frame $O$, respectively; $d_1$ and $d_2$ represent the linear displacements of the two prismatic joints; and $\mathbf{R}_y(\cdot)$ and $\mathbf{R}_z(\cdot)$ are the rotation matrices about the $y$- and $z$-axes, respectively.

\subsection{PIP Kinematics}

The PIP joint is connected to joint $P_3$ via a crossed four-bar linkage ($P_3$–$P_4$–$P_5$–$P_{pip}$, see Figure~\ref{fig:finger_kinematics}-c), while joint $P_3$ is actuated by the third motor through a PSU chain ($P_2$–$P_1$–$P_0$, see Figure~\ref{fig:finger_kinematics}-b).

\textbf{Crossed Four-bar Linkage} Frame $P_3$ is defined at joint $P_3$ by rotating an angle $\alpha$ about the $z$-axis of frame $P_{mcp}$ and translating by a position vector $\mathbf{P}_{P_3}^{P_{mcp}}$ from the origin of frame $P_{mcp}$. The relationship between $\alpha$ and the joint variable $q_3$ is expressed as $g_1(q_3, \alpha) = 0$. We analyze the kinematics of the crossed four-bar linkage (see Figure~\ref{fig:finger_kinematics}-c), and $g_1$ is derived as follows:
\begin{equation}\label{eq:q3_alpha}
\begin{gathered}
  \mathbf{R}^{P_{mcp}}_{P_{pip}} = \mathbf{R}_z(q_3),\\
  \mathbf{p}_{P_5}^{P_{mcp}} = \mathbf{p}_{P_{pip}}^{P_{mcp}}+ \mathbf{R}^{P_{mcp}}_{P_{pip}} \mathbf{p}_{P_5}^{P_{pip}},\\
  \mathbf{R}^{P_{mcp}}_{P_3} = \mathbf{R}_z(\alpha),\\
  \mathbf{p}_{P_4}^{P_{mcp}} = \mathbf{p}_{P_3}^{P_{mcp}}+ \mathbf{R}^{P_{mcp}}_{P_{3}} \mathbf{p}_{P_4}^{P_{3}},\\
 ||\mathbf{p}_{P_5}^{P_{mcp}} - \mathbf{p}_{P_4}^{P_{mcp}}|| = ||\overrightarrow{P_4P_5}||.\\
\end{gathered}
\end{equation}

\textbf{PSU Chain} The PSU chain (see Figure~\ref{fig:finger_kinematics}-b) is analyzed to derive the relationship between $\alpha$ and $d_3$, expressed as $g_2(\alpha, d_3) = 0$:
\begin{equation}\label{eq:alpha_d3}
\begin{gathered}
\mathbf{p}_{P_{22}}^{P_{mcp}} = \mathbf{p}_{P_3}^{P_{mcp}} + \mathbf{R}^{P_{mcp}}_{P_3} \mathbf{p}_{P_{22}}^{P_3},\\
\mathbf{p}_{P_{22}}^O = \mathbf{p}_{P_{mcp}}^O + \mathbf{R}^{O}_{P_{mcp}} \mathbf{p}_{P_{22}}^{P_{mcp}},\\
\mathbf{p}_{P_1}^O = \mathbf{p}_{P_{10}}^O - [d_3, 0, 0]^T,\\
||\mathbf{p}_{P_{22}}^O - \mathbf{p}_{P_1}^O|| = \|\overrightarrow{P_1P_{22}}\|^2 = \|\overrightarrow{P_1P_{21}}\|^2 + \|\overrightarrow{P_{21}P_{22}}\|^2 - 2 \|\overrightarrow{P_1P_{21}}\| \|\overrightarrow{P_{21}P_{22}}\| \cos(\pi - |q_1|),
\end{gathered}
\end{equation}
where $\mathbf{p}_{P_{10}}^O$ is the initial position vector of point $P_1$ expressed in frame $O$, and $d_3$ represents the linear displacements of the third prismatic joint.

Notably, we revise the representation of the universal joint at $P_2$, originally shown in Figure 3 of~\citep{KimILDA2021}, to accurately reflect its actual implementation (see the coral-shaded depiction in Figure~\ref{fig:finger_kinematics}-b). Additionally, we approximate the angle $q$—defined between vectors $\overrightarrow{P_{21}P_1}$ and $\overrightarrow{P_{22}P_{21}}$—by $q_1$, and apply the \textit{Law of Cosines} to formulate the final equation.

\subsection{DIP Kinematics}
The kinematics of a crossed four-bar linkage (see Figure~\ref{fig:finger_kinematics}-d) is analyzed to derive the relationship between $q_3$ and $q_4$, expressed as $h(q_3, q_4)=0$:
\begin{equation}\label{eq:alpha_d3}
\begin{gathered}
\mathbf{R}^{P_{pip}}_{P_{dip}} = \mathbf{R}_z(q_4),\\
\mathbf{p}_{P_{7}}^{P_{pip}} = \mathbf{p}_{P_{dip}}^{P_{pip}} + \mathbf{R}^{P_{pip}}_{P_{dip}} \mathbf{p}_{P_{7}}^{P_{dip}},\\
\mathbf{p}_{P_{6}}^{P_{mcp}} = \mathbf{p}_{P_{pip}}^{P_{mcp}} + \mathbf{R}^{P_{mcp}}_{P_{pip}} \mathbf{p}_{P_{6}}^{P_{pip}},\\
||\mathbf{p}_{P_{6}}^{P_{pip}} - \mathbf{p}_{P_{7}}^{P_{pip}}|| = ||\overrightarrow{P_6P_7}||.
\end{gathered}
\end{equation}

\end{document}